\documentclass[letterpaper, 10 pt, conference]{ieeeconf}  % Comment this line out if you need a4paper
\usepackage{amsmath,amsfonts}
\usepackage{algorithmic}
\usepackage{algorithm}
\usepackage{array}
\usepackage{booktabs}
\usepackage{subcaption}
\usepackage{textcomp}
\usepackage{stfloats}
\usepackage{url}
\usepackage{verbatim}
\usepackage{graphicx}
\usepackage{tabularx}
\usepackage{multirow}
\usepackage{multicol}
\usepackage{array}
\usepackage{makecell}              
\usepackage{booktabs}
\usepackage{utfsym}
\usepackage{color}
\usepackage{amssymb}
\usepackage{threeparttable}
\usepackage[colorlinks, 
            linkcolor=blue,   
            anchorcolor=blue,
            citecolor=green,  
            urlcolor=blue]{hyperref} 
\usepackage[nocompress]{cite} 
\usepackage{stfloats}
\usepackage{color}
\usepackage{ifthen}
\usepackage{fancyhdr}
\usepackage{float}
\usepackage{cuted}
\usepackage{orcidlink}

\IEEEoverridecommandlockouts                              % This command is only needed if 
\title{\LARGE \bf
DiffPlace: Street View Generation via Place-Controllable Diffusion Model Enhancing Place Recognition
}

\author{Ji Li$^{\orcidlink{0009-0009-7463-1343}}$, 
    Zhiwei Li$^{\orcidlink{0009-0006-5694-3026}}$,
    Shihao Li$^{\orcidlink{0000-0002-8094-8701}}$,
    Zhenjiang Yu$^{\orcidlink{0009-0000-6931-3230}}$,
    Boyang Wang$^{\orcidlink{0000-0003-3613-8792}}$
    and Haiou Liu$^{\orcidlink{0009-0001-8904-8101}}$ % <-this % stops a space
\thanks{The research is funded by the National Natural Science Foundation of China under Grants No. 52302489; and the National Natural Science Foundation of China under Grants No. 52172378. (Corresponding author: Boyang Wang)
}% <-this % stops a space
\thanks{Project page: 
\href{https://jerichoji.github.io/DiffPlace/}{https://jerichoji.github.io/DiffPlace/}}}
\begin{document}
\maketitle
\thispagestyle{empty}
\pagestyle{empty}

%%%%%%%%%%%%%%%%%%%%%%%%%%%%%%%%%%%%%%%%%%%%%%%%%%%%%%%%%%%%%%%%%%%%%%%%%%%%%%%%
\begin{abstract}

Generative models have advanced significantly in realistic image synthesis, with diffusion models excelling in quality and stability. Recent multi-view diffusion models improve 3D-aware street view generation, but they struggle to produce place-aware and background-consistent urban scenes from text, BEV maps, and object bounding boxes. This limits their effectiveness in generating realistic samples for place recognition tasks.
To address these challenges, we propose DiffPlace, a novel framework that introduces a place-ID controller to enable place-controllable multi-view image generation. The place-ID controller employs linear projection, perceiver transformer, and contrastive learning to map place-ID embeddings into a fixed CLIP space, allowing the model to synthesize images with consistent background buildings while flexibly modifying foreground objects and weather conditions.
Extensive experiments, including quantitative comparisons and augmented training evaluations, demonstrate that DiffPlace outperforms existing methods in both generation quality and training support for visual place recognition. Our results highlight the potential of generative models in enhancing scene-level and place-aware synthesis, providing a valuable approach for improving place recognition in autonomous driving.
\end{abstract}

%%%%%%%%%%%%%%%%%%%%%%%%%%%%%%%%%%%%%%%%%%%%%%%%%%%%%%%%%%%%%%%%%%%%%%%%%%%%%%%%
\section{Introduction}

In recent years, generative models \cite{VAE, GAN, SD} have seen significant progress, particularly in the generation of high-quality and realistic visual content. Diffusion models \cite{LD, SD}, one of the major contributors to this progress, are especially known for their stable and high-quality sample generation. Recent breakthroughs in controllable generative technologies have further enabled precise and flexible content customization. In particular, the development of multi-view diffusion models have greatly improved the synthesis of street images with 3D geometry control. Initial studies focused on generating street view images to enhance image-based bird's-eye view (BEV) perception methods, with models like BEVGen \cite{BEVGen}, BEVControl \cite{BEVControl}, and MagicDrive \cite{MagicDrive}. More recent research has expanded to generating driving scene videos \cite{DrivingDiffusion, DriveDreamer,StreetScapes,HoloDrive}, aiming to enhance various aspects of the scene 
\begin{figure}[htbp]
    \centering
        \subcaptionbox{\label{fig:1a}}{
            \includegraphics[width = 0.4\textwidth]{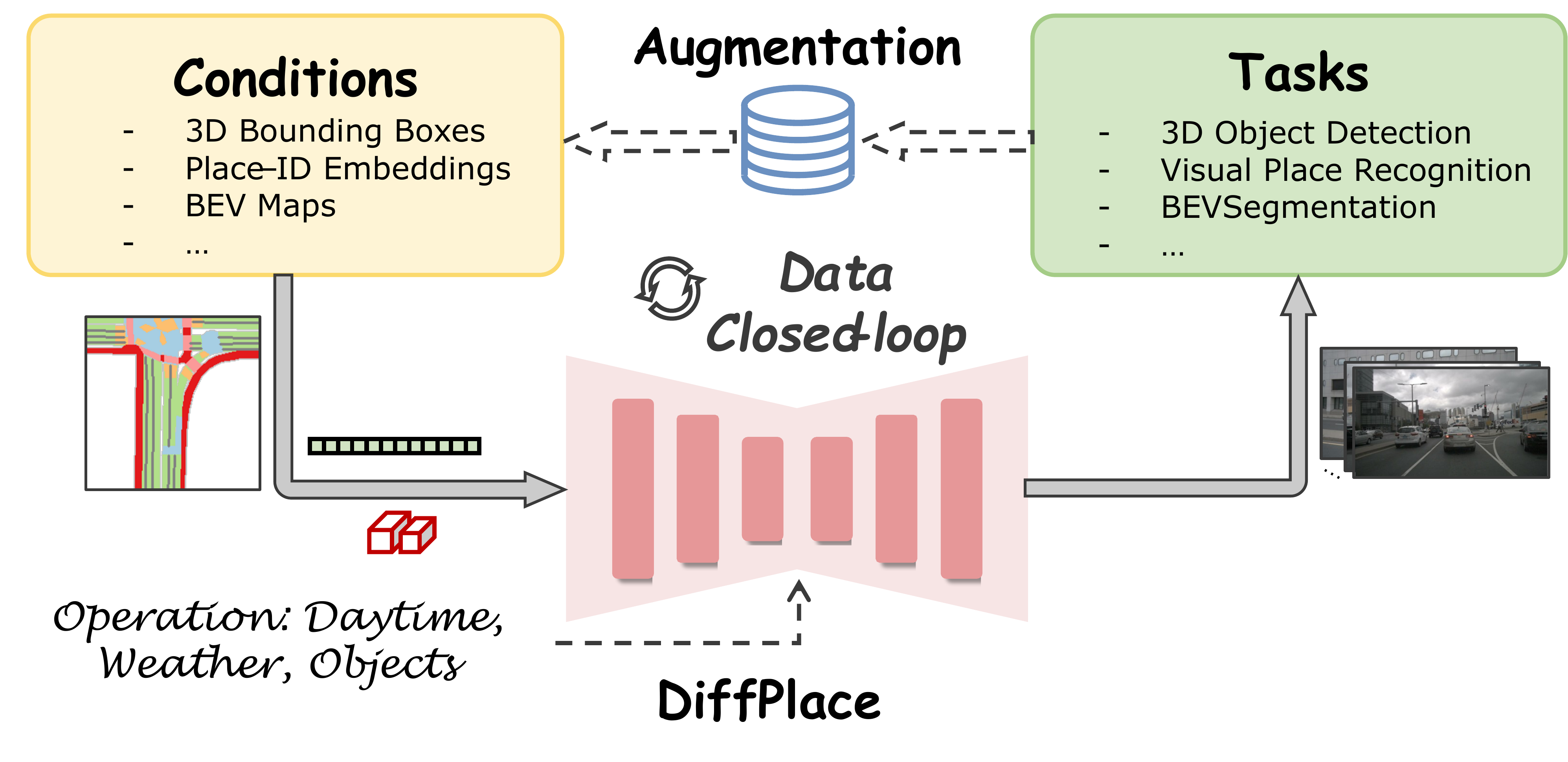}
            }
	\hfill
	\subcaptionbox{\label{fig:1b}}{
            \includegraphics[width = 0.4\textwidth]{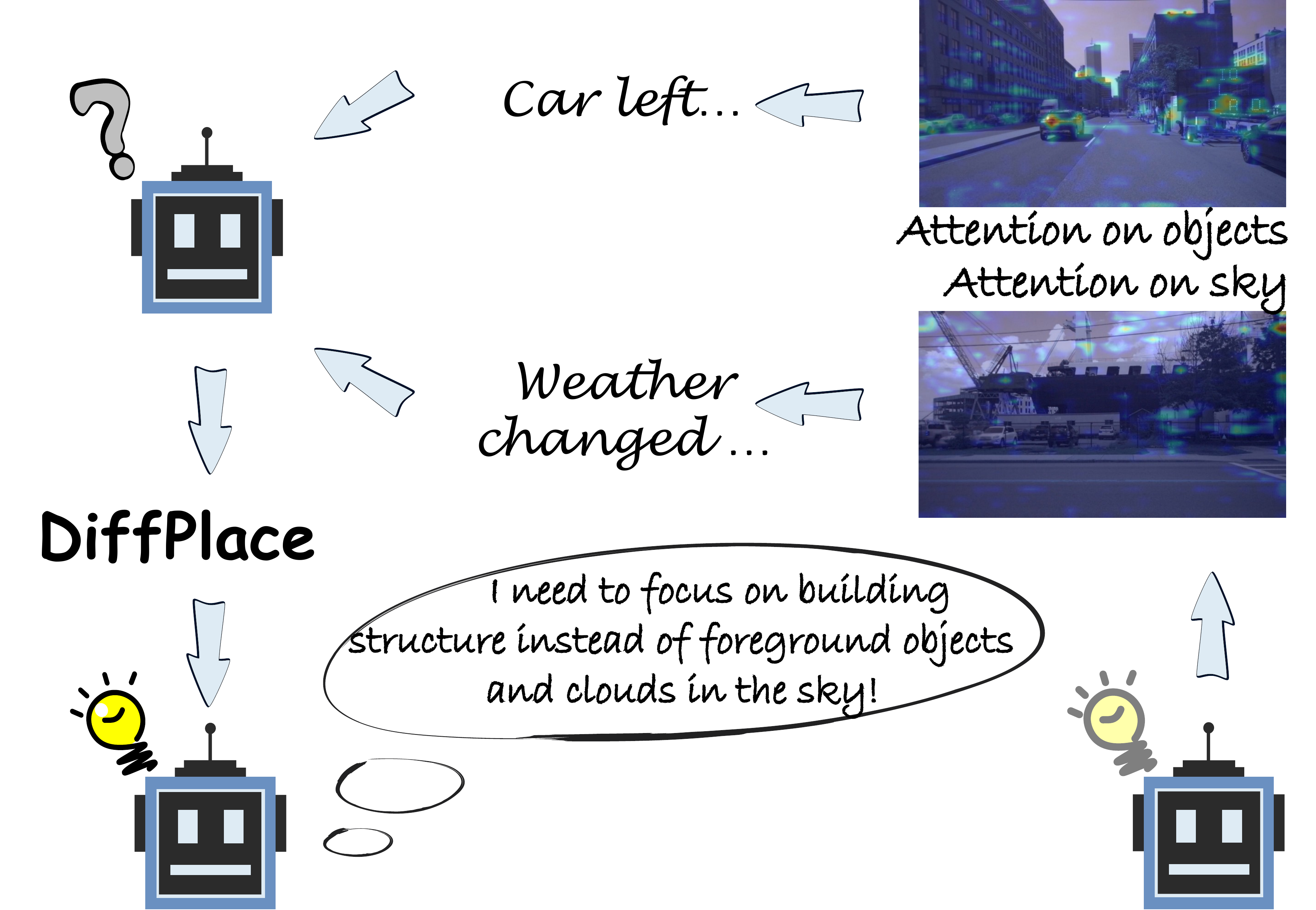}
            }
    \caption{\textbf{Systematic depiction of proposed DiffPlace.} (a) Generation is seen as the reverse process of perception, which generates images with the input of place-ID embeddings, bounding boxes, etc. We are the first to achieve data closed-loop for visual place recognition augmentation training. (b) The original place recognition model mistakenly focused on foreground objects and clouds in the sky. We corrected the place recognition model by augmented training on ``car left'' and ``weather changed'' situations through DiffPlace.}
    \label{fig:1}
    \vspace{-25pt}
\end{figure}
generation. However, generating background-consistent (or place-aware) street view images remains a challenge, as current methods rely solely on text, BEV maps, and object bounding boxes.

While these components are useful, they do not provide sufficient information for accurately modeling the background of a scene, limiting the realism of the generated content. As a result, these methods struggle to create synthetic samples that are closely resemble from real-world locations. 
This limitation restricts the use of generative models in place recognition tasks \cite{NetVLAD, RangePlace}, which remains one of the most important yet challenging tasks in autonomous driving and robotics.
% high discriminative accuracy is crucial
% With recent advancements in generative models, the question arises: can these innovations help push forward the progress of place recognition? 
A straightforward approach to address this might involve fine-tuning text-conditioned diffusion models directly on place-ID embeddings to enhance their place-aware prompting capabilities. However, this strategy has several drawbacks. First, it compromises the ability of street view generative models to generate images from text, BEV maps, and 3D bounding boxes. Additionally, fine-tuning such models demands substantial computational resources. To overcome these challenges, we propose a novel framework for street view generation to generate background-controllable multi-view images with additional place-ID embeddings. These embeddings integrate seamlessly with existing controllability.

Our work introduces \textbf{DiffPlace}, a method that leverages visual place recognition networks and a place-ID controller to enable place-aware scene synthesis. The core innovation of the place-ID controller involves linear projection, perceiver transformer, and contrastive learning to map place-ID embeddings into the fixed CLIP space. Through place-aware control, the synthesized images maintain consistent background information but allow modifications to foreground objects and weather conditions. These generated images ensure that the generated scenes retain key background elements while allowing for variability in foreground details, thus they can be used to augment place recognition training. We summarize our contributions as follows:

% \noindent\textbf{Summary of Contributions:}
\begin{itemize}
\item To the best of our knowledge, we are the first to propose a place-controllable diffusion model for generating augmented data to improve the performance of existing place recognition methods. Our approach enables fine-grained control over both objects-level and scene-level synthesis as verified by extensive quantitative experiments.
\item Our method generates high-fidelity street view images that preserve background consistency across varying weather conditions and different foreground objects by simply adjusting the prompts. Using our synthetic images as augmented data significantly enhances the training of place recognition models, demonstrating superior performance.
\end{itemize}

% With the place-ID controller, we can generate high-fidelity street view images that consistently reflect background details, even under varying weather conditions and different foreground objects, simply by adjusting the input prompts. This significantly enhances the performance of both place recognition and object detection tasks. We compare our method with existing approaches through quantitative side-by-side evaluations and augmentation experiments, demonstrating superior performance in both image generation quality and place recognition tasks.

\section{Related Work}
\subsection{Latent Diffusion Models}
The task of text-to-image (T2I) generation focuses on creating realistic images from textual descriptions. Diffusion models, a class of probabilistic generative models, introduce noise to data in a gradual manner and then learn to reverse this process in order to generate samples \cite{SD}. Early approaches framed this challenge as a sequence-to-sequence problem. In recent advancements, Denoising Diffusion Probabilistic Models (DDPM) \cite{LD} have demonstrated substantial success in addressing the T2I task. The image quality has been further enhanced by leveraging the strong image generation capabilities of diffusion models, or by improving the alignment between text and images through powerful text encoders \cite{CLIP}. For instance, DALL-E \cite{DALL-E} employs text tokens to generate discrete image embeddings via VQ-VAE. Further improvements have been made in subsequent works, utilizing more sophisticated architectures like encoder-decoder frameworks \cite{U-Net}, and hierarchical transformers \cite{DiT}.

These models have gained considerable attention in recent years due to their impressive performance across a variety of applications, setting new benchmarks in areas such as video generation \cite{VD3D}, and 3D content creation \cite{ImageDream}. In order to further improve the controllability of image generation, techniques like ControlNet \cite{ControlNet} and GLIGEN \cite{GLIGEN} have been introduced, enabling the use of various control signals such as depth maps, segmentation maps, canny edges, and sketches.
% SD CAT3D 4DiM

\subsection{Visual Place Recognition}

Visual Place Recognition (VPR) has progressed from traditional feature-based approaches, such as RootSIFT \cite{RootSIFT} to more advanced deep learning methods. Early CNN-based models \cite{NetVLAD} achieved notable results, but recently, Vision Transformers (ViTs) \cite{SALAD, R2Former} have emerged, providing enhanced performance owing to their capacity to grasp long-distance relationships. Visual Foundation Models (VFMs) like CLIP \cite{CLIP} and DINOv2 \cite{DINOv2} have gained popularity, with solutions such as AnyLoc \cite{AnyLoc} using dense local features for zero-shot VPR. Although these models perform well in certain contexts, they struggle when faced with significant time gaps or environmental changes. Fine-tuned models like SALAD \cite{SALAD} enhance accuracy but require increased feature dimensionality and higher memory consumption. Other approaches, such as CricaVPR \cite{CricaVPR}, integrate trainable adapters within ViT architectures to prevent catastrophic forgetting, though they still confront computational difficulties.
Through early efforts \cite{revise1,revise2} have explored GAN models for visual place recognition under changing environments, existing VPR methods remain challenged by changing foregrounds, lighting, and weather conditions. In this paper, we focus on using diffusion models to generate richer training data, incorporating diverse foreground objects and varying weather conditions. This approach aims to enhance the training of any existing VPR algorithms, and improve their robustness in real-world applications.

\subsection{Street View Image generation}
This task can be seen as the reverse of perception tasks, aiming to produce images from inputs like bounding boxes. Previous techniques employed GANs \cite{Spatial} or diffusion models \cite{RAL1, RAL2} to synthesize images based on 2D layouts. These approaches commonly encoded the layout into a conditional image, which was subsequently processed through downsampling and upsampling alongside the data. More contemporary work, such as \cite{BEVGen} has leveraged VQ-VAE to generate multi-view urban street view images from BEV layouts.

At the same time, approaches such as BEVControl \cite{BEVControl}, MagicDrive \cite{MagicDrive}, and DrivingDiffusion \cite{DrivingDiffusion} have incorporated layout specifications to further refine the image generation process. The essence of diffusion-based generative models resides in their capacity to grasp the complexities of the environment. By harnessing the potential of these models, \cite{SGD,MagicDrive3d} then introduce 3D Gaussian Splatting \cite{3DGS} representations to render novel-view images in generated scenes.
Building upon MagicDrive, we tackle a more challenging task of generating place-aware multi-view images. Then we treat the generated images as auxiliary datasets for visual place recognition.

\begin{figure}[htbp]
	\center{\includegraphics[width=\linewidth]  {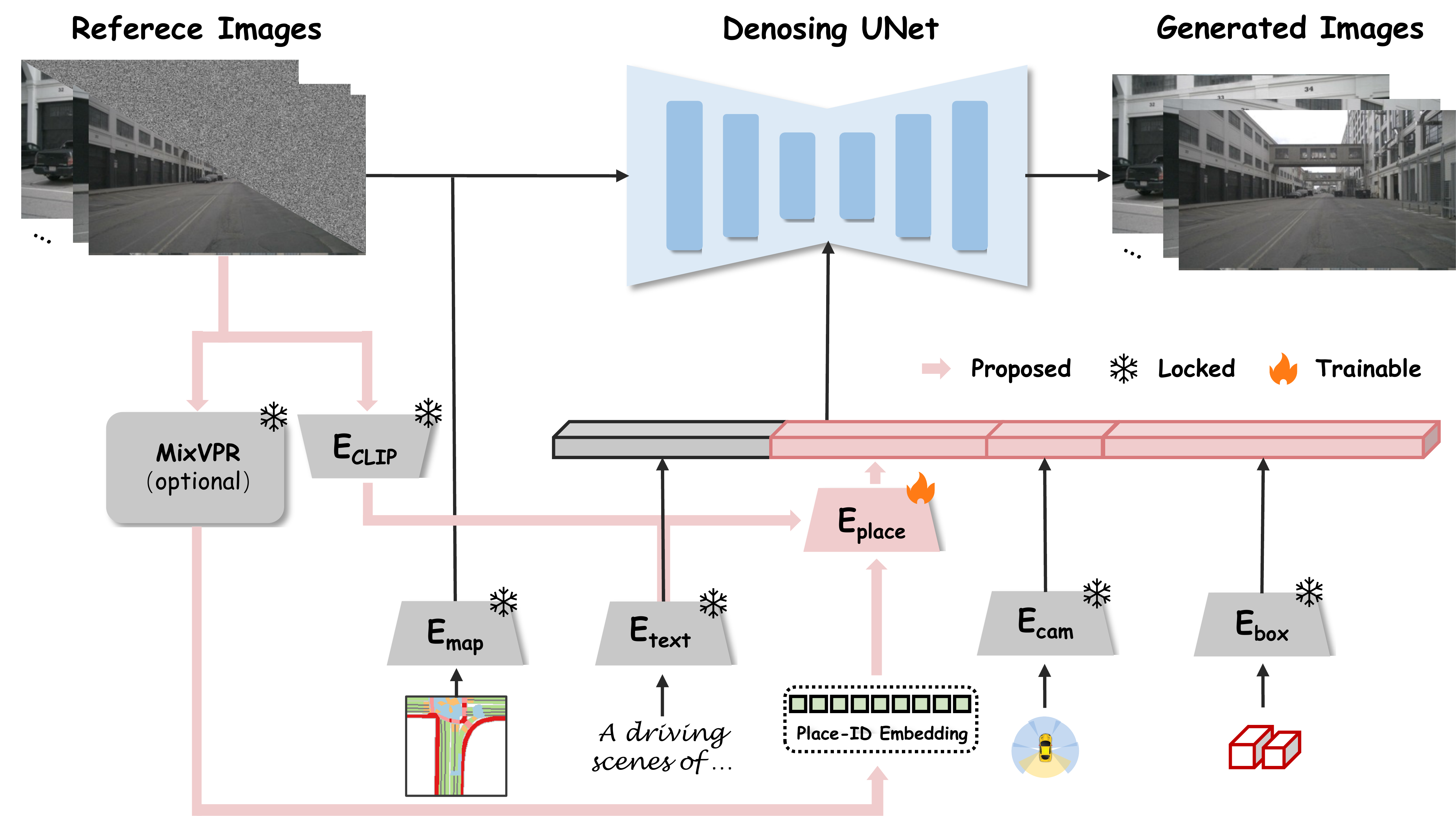}} 
	\caption{\textbf{Overview of the DiffPlace pipeline.} The input scene representation \( S = \{ \textit{\text{MAP}, \text{BOX}, \text{TEXT}, \text{PLACE\_ID}} \} \) is processed by dedicated encoders: \( E_{\text{map}} \), \( E_{\text{text}} \), \( E_{\text{place}} \), \( E_{\text{cam}} \), and \( E_{\text{box}} \). The resulting encoded features are concatenated and fed into the U-Net via cross-attention mechanisms to generate multi-view consistent images with controllable background and foreground elements. An optional visual place recognition network (MixVPR \cite{MixVPR}) is utilized to extract place-ID embeddings, enabling enhanced place-aware synthesis.}
	\label{fig:2}
    \vspace{-10pt}
\end{figure}

\section{Methods}
While advanced generative methods can create high-quality images of driving scenes, their impact on downstream perception tasks such as place recognition (shown in Fig. \ref{fig:1a}), remains limited. We believe this is mainly due to the inadequate control over the generated background information, which is vital for effective scene understanding and place recognition.
As depicted in Fig. \ref{fig:2}, various strategies are implemented to inject information into multi-view diffusion models. We propose a place-ID controller that maps place-ID embedding from the place recognition network to align with the CLIP image space. Key components of this approach include place-ID encoding, attribute perceiver transformer, and contrastive learning strategy.

In Section 3-A, we begin by presenting the basic notions of multi-view diffusion. Subsequently, in Section 3-B, we unveil our comprehensive diffusion architecture, which incorporates bespoke designs aimed at bolstering the place-controllability of the diffusion models. Lastly, we provide a detailed implementation for adding control of place features to the diffusion model.

\subsection{Preliminary}

Latent Diffusion models are intended to capture a probability distribution, denoted as \( p_\theta(x_0) = p_\theta(x_{0:T}) \, dx_{1:T} \), where \( x_0 \) signifies the data and \( x_{1:T} := x_1, \ldots, x_T \) represent latent variables. This joint distribution is defined by a Markov chain \cite{LD}, specifically referred to as the reverse process:
\begin{equation}
p_\theta(x_{0:T}) = p(x_T) \prod_{t=1}^{T} p_\theta(x_{t-1} | x_t),
\label{eq:1}
\end{equation}
% \[
% p_\theta(x_{0:T}) = p(x_T) \prod_{t=1}^{T} p_\theta(x_{t-1} | x_t)
% \]
with \( p(x_T) = \mathcal{N}(x_T; 0, I) \) and \( p_\theta(x_{t-1} | x_t) = \mathcal{N}(x_{t-1}; \mu_\theta(x_t, t), \sigma^2_t I) \). Here, \( \mu_\theta(x_t, t) \) is a trainable component, while the variance \( \sigma^2_t \) consists of untrained time-dependent constants. The aim is to learn \( \mu_\theta \) for generation purposes.

To achieve this, the forward process is constructed:
% \[
% q(x_{1:T} | x_0) = \prod_{t=1}^{T} q(x_t | x_{t-1}),
% \]
\begin{equation}
q(x_{1:T} | x_0) = \prod_{t=1}^{T} q(x_t | x_{t-1}),
\label{eq:2}
\end{equation}
where 
\begin{equation}
q(x_t | x_{t-1}) = \mathcal{N}(x_t; \sqrt{1 - \beta_t} x_{t-1}, \beta_t I),
\label{eq:3}
\end{equation}
% \[
% q(x_t | x_{t-1}) = \mathcal{N}(x_t; \sqrt{1 - \beta_t} x_{t-1}, \beta_t I),
% \]
and \( \beta_t \) are constants. The DDPM approach demonstrates that by defining:
\begin{equation}
\mu_\theta(x_t, t) = \frac{1}{\sqrt{\alpha_t}} x_t - \beta_t \sqrt{1 - \bar{\alpha}_t} \epsilon_\theta(x_t, t),
\label{eq:4}
\end{equation}
% \[
% \mu_\theta(x_t, t) = \frac{1}{\sqrt{\alpha_t}} x_t - \beta_t \sqrt{1 - \bar{\alpha}_t} \epsilon_\theta(x_t, t),
% \]
with \( \alpha_t \) and \( \bar{\alpha}_t \) being constants derived from \( \beta_t \) and \( \epsilon_\theta \) functioning as a noise predictor, we can learn \( \epsilon_\theta \) by minimizing the following loss function:
\begin{equation}
\mathcal{L}_{\text{base}} = \mathbb{E}_{t, x_0, \epsilon} \left\| \epsilon - \epsilon_\theta\left(\sqrt{\bar{\alpha}_t} x_0 + \sqrt{1 - \bar{\alpha}_t} \epsilon, t\right) \right\|^2,
\label{eq:5}
\end{equation}
% \[
% \mathcal{L} = \mathbb{E}_{t, x_0, \epsilon} \left\| \epsilon - \epsilon_\theta\left(\sqrt{\bar{\alpha}_t} x_0 + \sqrt{1 - \bar{\alpha}_t} \epsilon, t\right) \right\|^2,
% \]
where \( \epsilon \) is a random variable sampled from \( \mathcal{N}(0, I) \).

To maintain viewpoint consistency, multi-view diffusion models, such as those proposed in \cite{MVDream,MVDiffusion++}, typically utilize cross-view attention modules. As expressed in Eq. \ref{eq:6}, given the sparse camera layout in driving environments, each cross-view attention mechanism enables the target view in accessing information from its neighboring left and right views, The target view states then consolidates this information through a skip connection. The cross-view attention \( \text{Attention}_{cv}\) computation can be described as follows:
\begin{equation}
\text{Attention}_{cv}(Q_t, K_i, V_i) = \text{softmax}\left(\frac{Q_t K_i^T}{\sqrt{d}}\right) \cdot V_i, \quad i \in \{l, r\}
\label{eq:6}
\end{equation}
In this context, \( t \), \( l \), and \( r \) refer to the target view, left view, and right view, respectively.

\subsection{Overall Architecture}

Our latent multi-view diffusion model consists of a VAE encoder \(E\), a denoising U-Net, and a VAE decoder \(D\). While description text, BEV maps, and 3D geometric information similar to MagicDrive \cite{MagicDrive} provide useful context, they do not offer precise guidance for generating the background. To address this, we introduce an additional controller \(E_{place}\), for place-ID, which expands the original input-output pair. Let \( S = \{ \textit{\text{MAP}, \text{BOX}, \text{TEXT}, \text{PLACE\_ID}} \} \) represent the components of a driving scene encompassing the ego vehicle, where \( \textit{\text{MAP}} \) is a binary map depicting a \( w \times h \) meter area of the road in Bird's Eye View (BEV), with \( c \) semantic classes. \( \textit{\text{BOX}} = \{(c_i, b_i)\}_{i=1}^{N} \) indicates the locations of 3D bounding boxes for each object within the scene. At the scene level, \( \textit{\text{TEXT}} \) includes textual descriptions offering fundamental context about the scene (such as weather and time of day). Additionally, \( \textit{\text{PLACE\_ID}} \) offers more detailed background information of the scene. Given the camera pose \( P = [K, R, T] \) (which includes intrinsic parameters, rotation, and translation), the generator \( G(\cdot) \) aims to synthesize multi-view consistent images that incorporate foreground (object-level), midground (map-level), and background (scene-level) information.

\subsection{Place-ID Encoding}

Place recognition networks are specifically designed to extract distinctive features of a given place. In our framework, we can leverage any existing visual place recognition method. We optionally take MixVPR \cite{MixVPR} as example, and adopt an implementation that utilizes a ResNet-50 backbone in combination with all-MLPs aggregation, producing a discriminative place-ID embedding with a dimension of 4096.

To ensure the dimension of the place-ID embedding aligns with other conditions, we implement two trainable linear projection layers. These layers operate on the place-ID embedding and output a sequence of features \( Z \) with length \( N_S \) (set to 4 in our implementation), matching the dimensionality of other prompt embeddings. It has been observed that using two linear layers results in better performance compared to employing an MLP with successive blocks \cite{IP-Adapter}.

We do not apply a masking mechanism to the place-ID embedding, as the aggregation of various scenes consistently yields a fixed-length place-ID embedding, thereby eliminating the need for random masking operations. Moreover, we refrain from using a multi-view approach in place recognition, as we believe it could disrupt the integrity of place information. Instead, we rely on camera parameter encoding and cross-view attention in the U-Net to maintain consistency across multiple views during the generating phase.

\subsection{Attribute Perceiver Transformer}
We employ a perceiver-based transformer to map the embedding with the aid of attributes in reference images. Prior to attribute extraction using the CLIP Image Encoder, we mask out the foreground objects using bounding boxes and the sky regions following the method in \cite{GroundedSAM}, thereby directing the network's focus to the street scene and minimizing interference from the other conditions. This approach ensures that the attention mechanism is concentrated solely on the relevant portions of the scene.

After being processed through a linear projector, the place-ID embeddings are enhanced by the CLIP image features from reference images through several cross-attention layers (we implement 3 layers with a dimension of 1024 in this study). Given the query place-ID embeddings \( Z \) and the CLIP image features \( c_I \), the output of the cross-attention mechanism is given by the following expression:

\begin{equation}
Z = \text{Attention}(Q, K, V) = \text{Softmax}\left(\frac{Q K^T}{\sqrt{d}}\right) V
\label{eq:7}
\end{equation}
where \( Q = Z W_q \) represents the query matrix for the place-ID embeddings, and \( K = c_I W_k \), \( V = c_I W_v \) denote the key and value matrices derived from the CLIP hidden states.

\begin{figure}[htbp]
	\center{\includegraphics[width=7cm]  {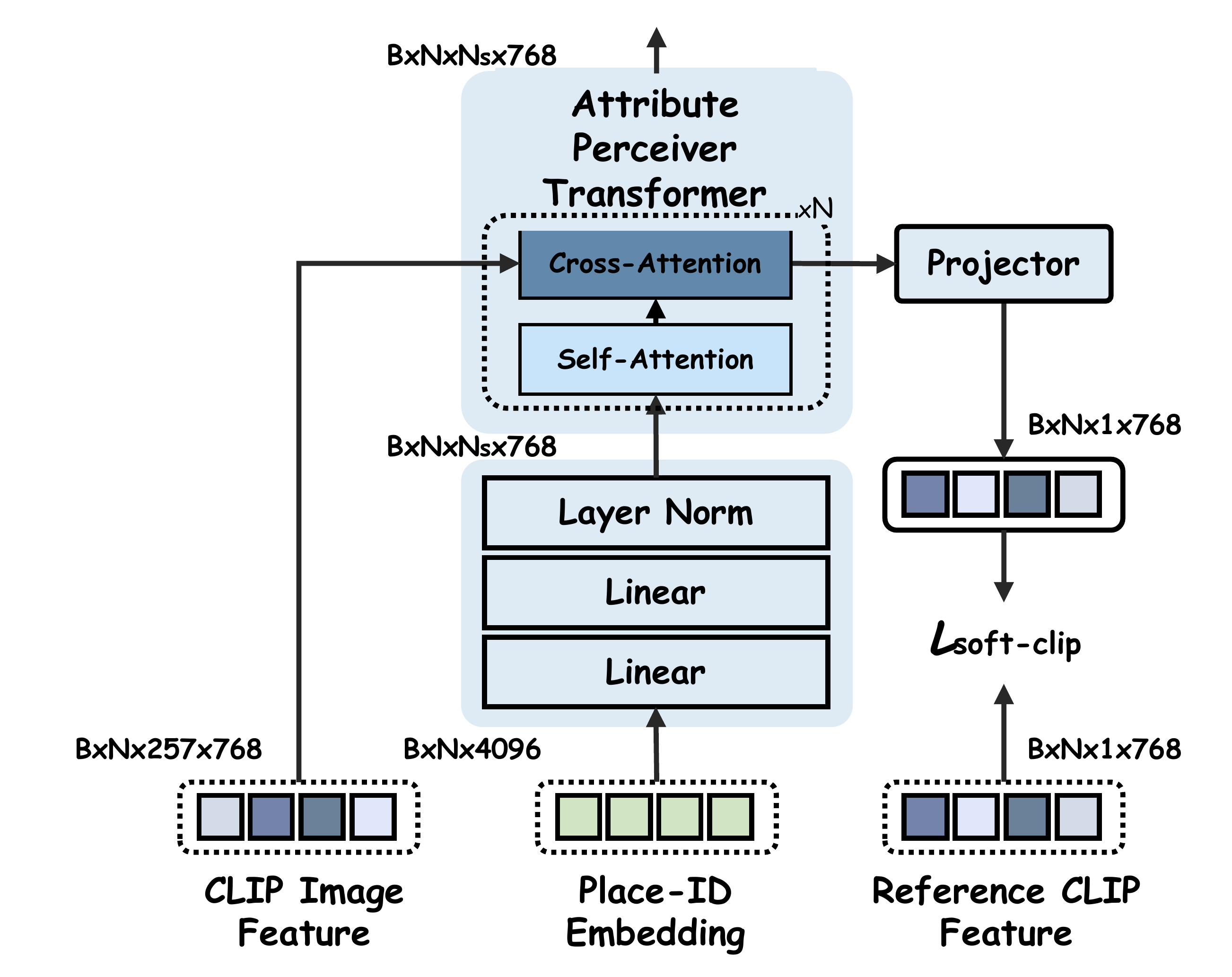}} 
	\caption{\textbf{Details of the proposed place-ID controller.} (a) Place-ID embeddings are projected via trainable linear layers to align with other conditions; (b) Attribute perceiver transformer interacts place-ID embeddings \( Z \)  with CLIP image features \( c_I \); (c) A contrastive loss $\mathcal{L}_{\text{SoftCLIP}}$ is applied to align place-ID embeddings with the CLIP latent space.}
	\label{fig:3}
        \vspace{-20pt}
\end{figure}

\subsection{Contrastive Learning}
We also integrate contrastive learning into the ControlNet training process, which aids in mapping place-ID embeddings into the CLIP space, thereby facilitating the alignment of place-ID conditions with the original text descriptions.

Contrastive learning is a potent technique for learning representations across different modalities by maximizing the cosine similarity of positive pairs while minimizing it for negative pairs. Previous studies have shown the efficacy of contrastive learning, even in the context of neural data‌ \cite{MindEye}. In our method, we apply contrastive learning to align additional place-ID embedding conditions with the static CLIP image space.
Specifically, we use a projector to transform features with a shape of \(N \times C\) to match the length of CLIP image tokens (after pooling in our implementation). These tokens are then normalized, and the SoftCLIP loss is computed between them. It is important to note that this strategy is only used during the training phase.

Our SoftCLIP loss is inspired by knowledge distillation \cite{Distilling}, which suggests that the softmax probability distribution generated by a strong teacher model provides a more effective teaching signal for the student model than hard labels. To create soft labels, we first calculate the dot product of the CLIP embeddings within a batch. The SoftCLIP loss is then computed as our contrastive objective between the CLIP-to-CLIP and place-ID-to-CLIP matrices, as follows:
\begin{align}
    \mathcal{L}_{\text{SoftCLIP}} & =  \sum_{i=1}^{N} \sum_{j=1}^{N}  \\
    & \quad \left[ \frac{\exp\left(\frac{t_i \cdot t_j}{\tau}\right)}{\sum_{m=1}^{N} \exp\left(\frac{t_i \cdot t_m}{\tau}\right)} \cdot \log\left( \frac{\exp\left(\frac{p_i \cdot t_j}{\tau}\right)}{\sum_{m=1}^{N} \exp\left(\frac{p_i \cdot t_m}{\tau}\right)} \right) \right]  \nonumber 
    \label{eq:8}
\end{align}
Therefore, the total loss is:
\begin{equation}
\mathcal{L} = \mathcal{L}_{\text{base}} + \lambda \cdot \mathcal{L}_{\text{SoftCLIP}}
\label{eq:9}
\end{equation}
where \( \lambda = 0.1 \).

\begin{figure*}[htbp]
	\center{\includegraphics[width=\linewidth]  {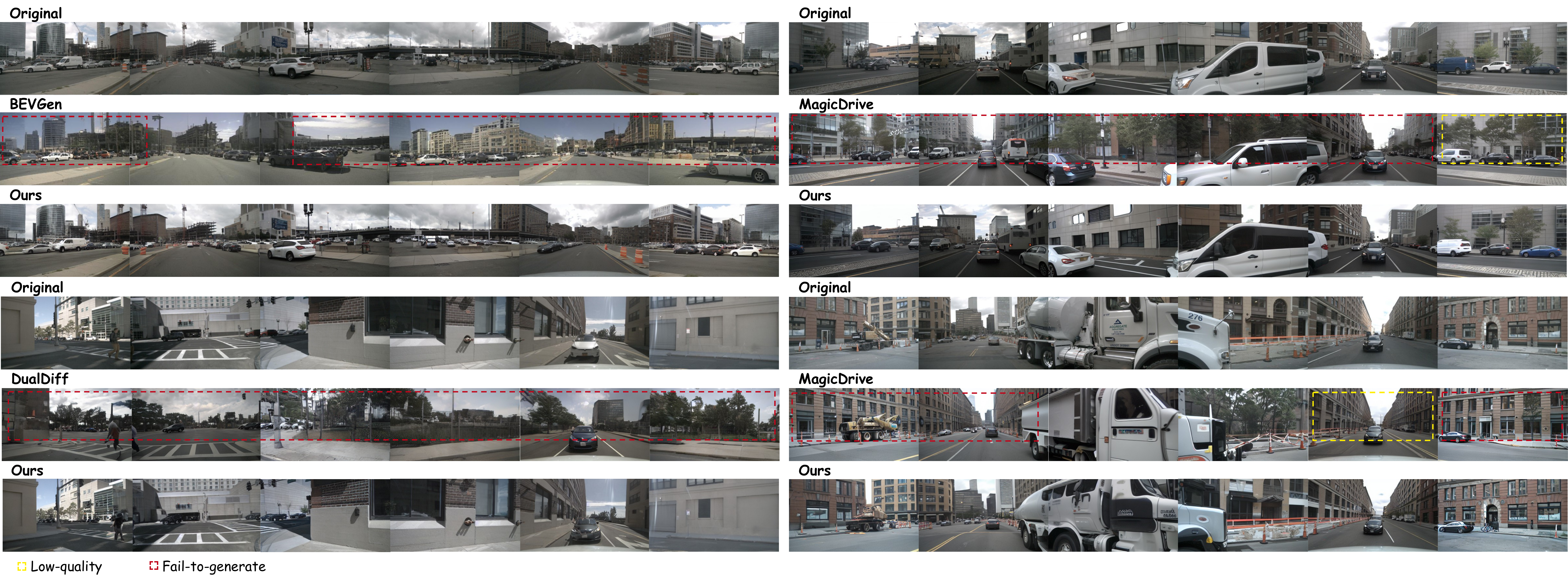}} 
	\caption{\textbf{Realism and controllability validation.} Our method demonstrates significantly better control over place features, particularly in the background, compared to BEVGen, MagicDrive and DualDiff. We highlight some background areas in low-quality (yellow) and fail-to-generate (red) for comparison. All scenes are from the nuScenes validation set.}
	\label{fig:4}
        \vspace{-20pt}
\end{figure*}

\section{Experiments}
We experimentally showcase the effectiveness of DiffPlace in enhancing place-aware controllability while preserving objects and weather controllability. Our approach achieves state-of-the-art performance in augmented training support for 3D object detection and place recognition benchmarks. We conduct extensive experiments and analyses to validate our design.

\subsection{Experiments Setups}

\subsubsection{Dataset and Baselines}
We conduct experiments on the nuScenes dataset to assess the effectiveness of DiffPlace. This dataset is widely used in 3D object detection and place recognition for autonomous driving. We follow standard methods for generating street view images, utilizing 700 street view scenes for training and 150 for validation. Moreover, our baselines are BEVGen\cite{BEVGen}, BEVControl\cite{BEVControl}, MagicDrive \cite{MagicDrive} and \cite{DualDiff}, the state-of-the-art approaches for controllable street view generation.

Additionally, we present training support for place recognition experiments on the Pitts30k-test dataset \cite{Pitts30k},which comprises images sourced from Google Street View and encompasses 8,000 database and 8,000 queries. The Pittsburgh dataset poses notable challenges owing to variations in viewpoint and lighting conditions. To examine the training support, synthesis images are generated from the training set of nuScenes dataset as augmentation data for training 3D object detection and place recognition models.

\subsubsection{Evaluation Metrics}
We assess the realism of the generated images using the Fr\`echet Inception Distance (FID). The images are generated in alignment with the validation set annotations, and we utilize pre-trained detection and recognition models on generated data to evaluate both image quality and control accuracy. To evaluate object generation accuracy controlled by the input 3D bounding boxes, we use the metrics such as mean Average Precision (mAP) and nuScenes Detection Score (NDS) in \cite{BEVFusion}. For place controllability, we adopt the same evaluation metrics as \cite{MixVPR}, where the Average Recall @1 and @5 (AR@1 and AR@5) are calculated. The criteria for considering a query image as successfully retrieved are met when at least one of the top-ranked reference images, either the top-1 or within the top-5, is situated within a 25-meter proximity to the query image.

% \begin{table*}[htbp]
% 	\caption{\textbf{ Comparison of generation fidelity with street view generation methods.} Conditions for data synthesis are from nuScenes validation set, which are not seen. For each task, we test the performance of corresponding models trained on the nuScenes training set.}
%         \label{tab:1} 
% 	\centering
% 	\belowrulesep=0pt
% 	\aboverulesep=0pt
% 	\renewcommand\arraystretch {1.5}
% 	\begin{threeparttable}       
% 		\begin{tabularx}{0.6\linewidth}{m{3cm}X|X<{\centering} X<{\centering}|X<{\centering} X<{\centering}}
% 			\toprule[1pt]
% 			\multirow{2}{*}{Method}  &
% 			\multirow{2}{*}{FID↓}  &
% 			\multicolumn{2}{c|}{3D Object Detection} &
% 			\multicolumn{2}{c}{Place Recognition} \\
% 			\cmidrule(lr{2pt}){3-4} \cmidrule(l{2pt}){5-6}
% 			&  & mAP↑ & NDS↑ & AR@1↑ & AR@5↑\\
% 			\midrule
% 			MagicDrive \cite{MagicDrive} & 16.2 & 12.3 &\textbf{ 23.32} & 35.9 & 64.1 \\
% 			DiffPlace (Ours) & \textbf{13.4} & \textbf{13.5} & 23.07  & \textbf{57.6}& \textbf{75.4} \\
% 			\bottomrule[1pt]
% 		\end{tabularx}
% 	\end{threeparttable}     
%     \vspace{-20pt}
% \end{table*}

\begin{table}[htbp]
\vspace{-10pt}
	\caption{\textbf{ Comparison of generation fidelity with street view generation methods.} On the generated images, We test the performance of MixVPR trained on the nuScenes training set. Conditions for data synthesis are from nuScenes validation set, which are not seen.}
        \label{tab:1} 
	\centering
	\belowrulesep=0pt
	\aboverulesep=0pt
	\renewcommand\arraystretch {1.5}
	\begin{threeparttable}       
		\begin{tabularx}{0.95\linewidth}{m{2cm}X|X<{\centering} X<{\centering}X<{\centering}}
			\toprule[1pt]
			\multirow{2}{*}{Method}  &
			\multirow{2}{*}{FID↓}  &
			\multicolumn{2}{c}{Place Recognition} &
                \multirow{2}{*}{Reference} \\
			\cmidrule(lr{2pt}){3-4}
			&  & AR@1↑ & AR@5↑\\
			\midrule
                BEVGen \cite{BEVGen} & 25.6  & 31.2 & 60.8 & RAL2024\\ 
                BEVControl \cite{BEVControl} & 24.8  & - & - & ICLR2024 \\
			MagicDrive \cite{MagicDrive} & 16.2  & 35.9 & 64.1 & ICLR2024 \\
                DualDiff \cite{DualDiff} & \textbf{11.0}  & 48.7 & 68.9 & Arxiv2025 \\
			DiffPlace (Ours) & 13.4  & \textbf{57.6} & \textbf{75.4} & - \\
			\bottomrule[1pt]
		\end{tabularx}
	\end{threeparttable}     
    \vspace{-10pt}
\end{table}

\subsubsection{Implementation Details}
We use the CLIP ViT-L/14 \cite{CLIP} as a frozen image encoder to extract attribute features in the place-ID controller. Experiments are time-consuming as the driving scenes involve 6 different views. Therefore, our DiffPlace leverages pre-trained weights from MagicDrive, which is implemented using ControlNet and Stable Diffusion v1.5, to reduce training costs. For the place recognition task, we trained a MixVPR model on the nuScenes training set to generate original place-ID embeddings, following the hyperparameter settings from \cite{MixVPR}. In training support experiment, we train this model on the same training set, augmented with generated data.

During training, we only optimize the place-ID controller while keeping the original fixed. The model is trained with a constant learning rate of \(1 \times 10^{-4}\) and a linear warm-up for the first 3000 iterations. The training process is conducted on 4 NVIDIA 3090 GPUs with a gradient descent batch size of 24. We use AdamW as the optimizer, with a weight decay of 0.01. To reconcile discrepancies in the object detection task, we use a resolution of \(224 \times 400\). For place recognition, we resize the images to \(320 \times 320\) to meet the required input size. For image reconstruction, we use 20 denoising timesteps with the UniPC \cite{UniPC} multi-step noise scheduler.

\vspace{-5pt}
\subsection{Quantitative Results}
\subsubsection{Realism and Controllability Validation}
To validate the effectiveness of DiffPlace in producing realistic imagery, we utilize the nuScenes validation dataset to synthesize street view images and present the relevant metrics in Table \ref{tab:1}. DiffPlace outperforms the baseline approach in terms of image fidelity, achieving notably lower FID scores. Although there exists an inherent discrepancy between place-ID embeddings and pre-trained models, our refinement technique successfully bridges this gap, resulting in highly realistic and convincing outputs.
Regarding controllability, while maintaining the same effect in the 3D object detection task, DiffPlace dramatically exceeds baseline results in place recognition task. This is attributed to the distinct place-ID controller which significantly boosts generation precision on background buildings.

As shown in Figure~\ref{fig:4}, our method is capable of generating background buildings that closely resemble those in the ground truth. Furthermore, Table~\ref{tab:1} demonstrates that our generated images achieve significantly higher average recall on the validation set—outperforming MagicDrive by 21.7\% and 11.3\%, and exceeding DualDiff by 8.9\% and 7.6\%, respectively. Compared with DualDiff, although it produces images with higher photorealism, the recognition results from MixVPR validate the effectiveness of our background (scene-level) controllability, as our generated images are more easily recognized by the pre-trained model. Notably, the validation set was not seen during training.

\subsubsection{Visualization Experiments}
\begin{figure}[htbp]
	\centering
	\begin{minipage}{0.3\linewidth}
		\centering
		\includegraphics[width=\linewidth]{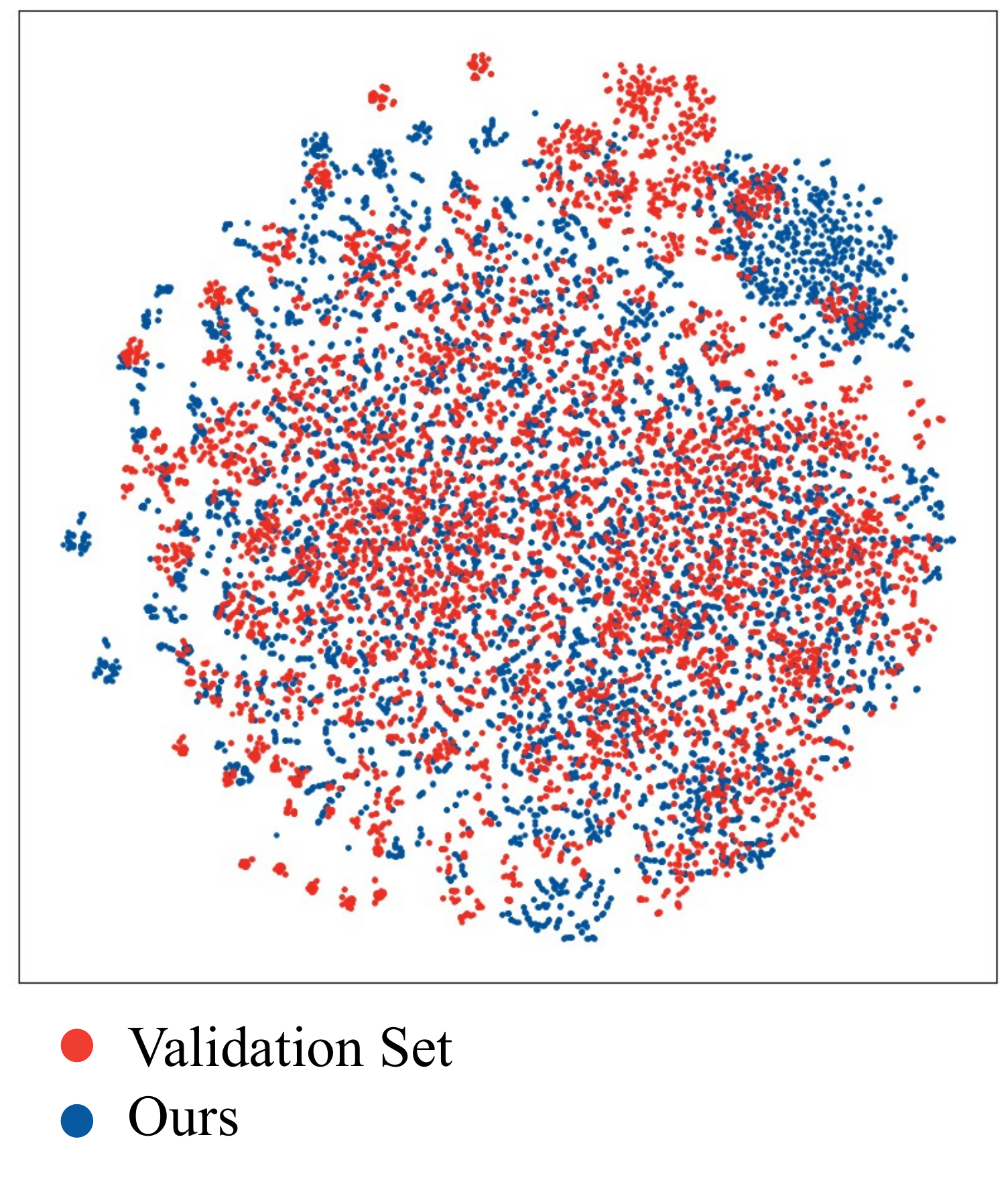}
            \vspace{-20pt}
            \label{fig5a}
		\caption*{(a)}
	\end{minipage}
        \hfill
	\begin{minipage}{0.3\linewidth}
		\centering
		\includegraphics[width=\linewidth]{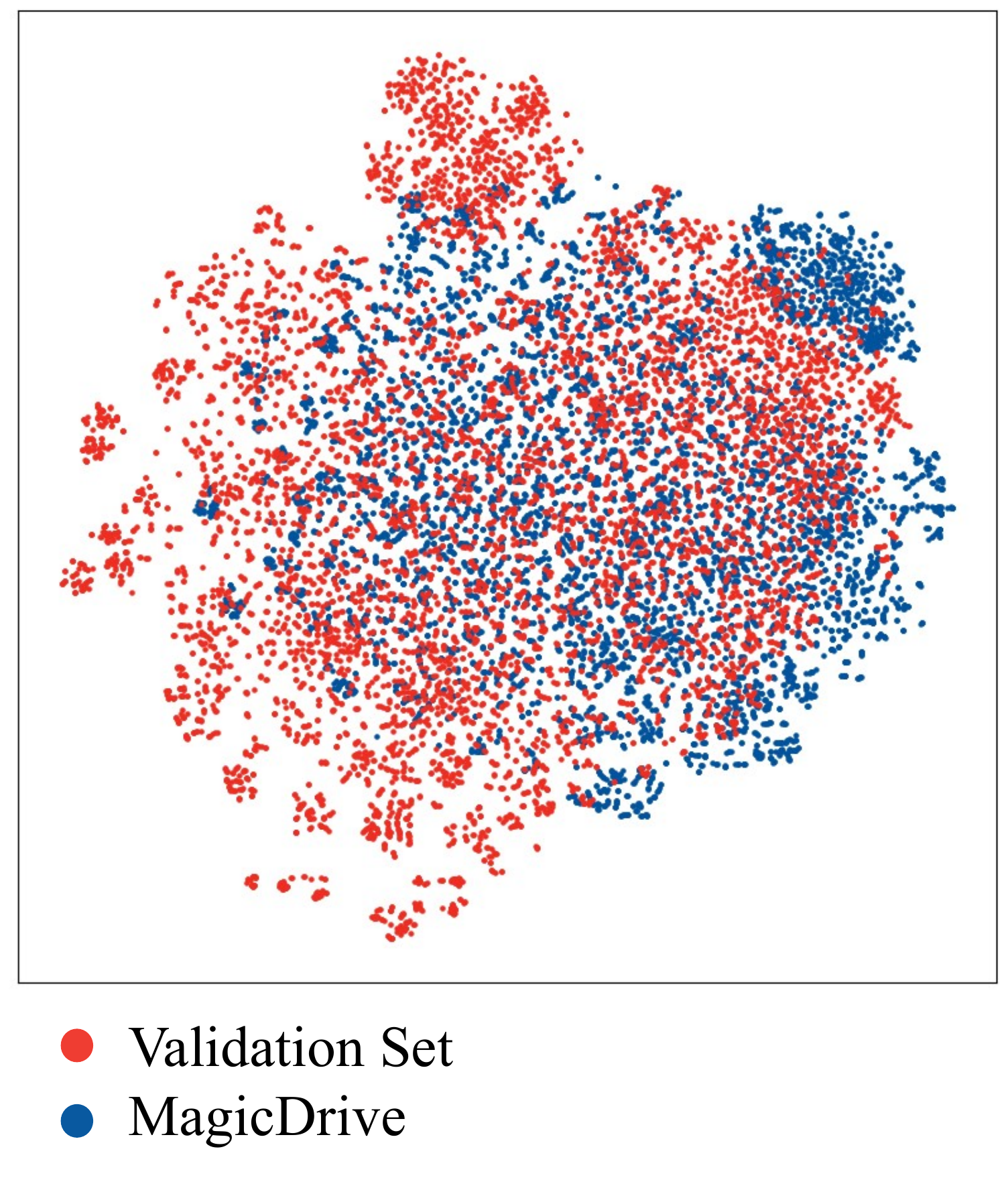}
            \vspace{-20pt}
            \label{fig5b}
		\caption*{(b)}
	\end{minipage}
	\hfill
	\begin{minipage}{0.34\linewidth}
		\centering
		\includegraphics[width=\linewidth]{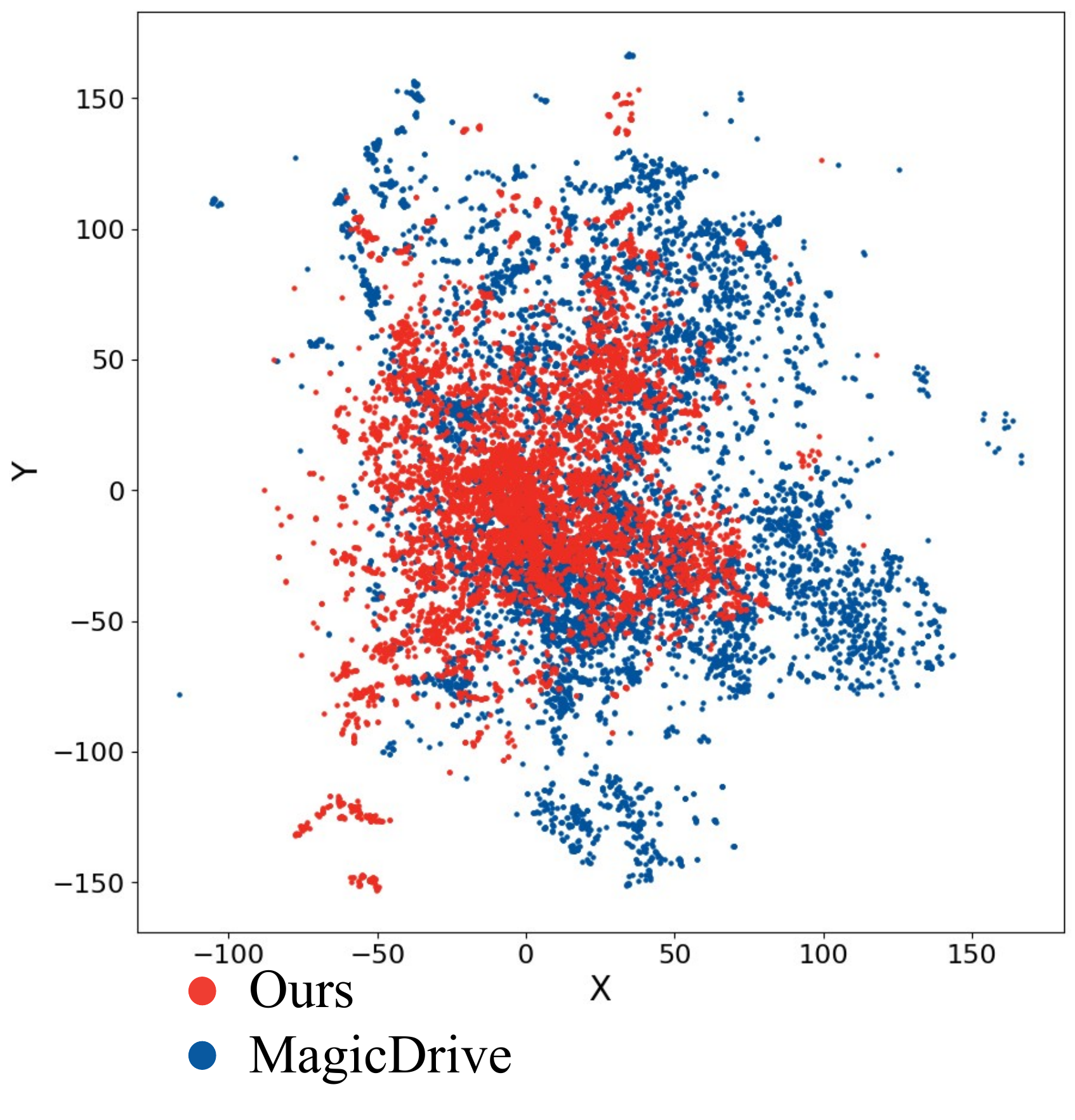}
            \vspace{-20pt}
            \label{fig5c}
		\caption*{(c)}
	\end{minipage}
    
    \caption{\textbf{Descriptors visualization results.} (a) and (b) are t-SNE Visualization of place-ID Embedding between nuScenes validation set and synthetic images of ours and MagicDrive's; (c) Euclidean Distance of place-ID Embedding between nuScenes validation set and synthetic images.}
    \label{fig:5}
\end{figure}

In addition, we utilize the t-SNE and Euclidean Distance of the place-ID embedding (in the shape of $4096$) generated by MixVPR on the original nuScenes validation set and synthetic images to conduct visualization experiments. As shown in Fig. \ref{fig:5}a and Fig. \ref{fig:5}b, compared with MagicDrive, the t-SNE results of ours show a smaller gap, which indicates the images generated by our method can generate images with more-aligned place feature with original ones. Then, the Euclidean Distance between original and synthetic ones in Fig. \ref{fig:5}c also revealed that our method has better controllability on place features. Additionally, we extract the attention scores within the cross-attention mechanism influenced by the place-ID embedding to visualize its role during the generation process.

\begin{figure}[htbp]
\vspace{-10pt}
	\center{\includegraphics[width=\linewidth]  {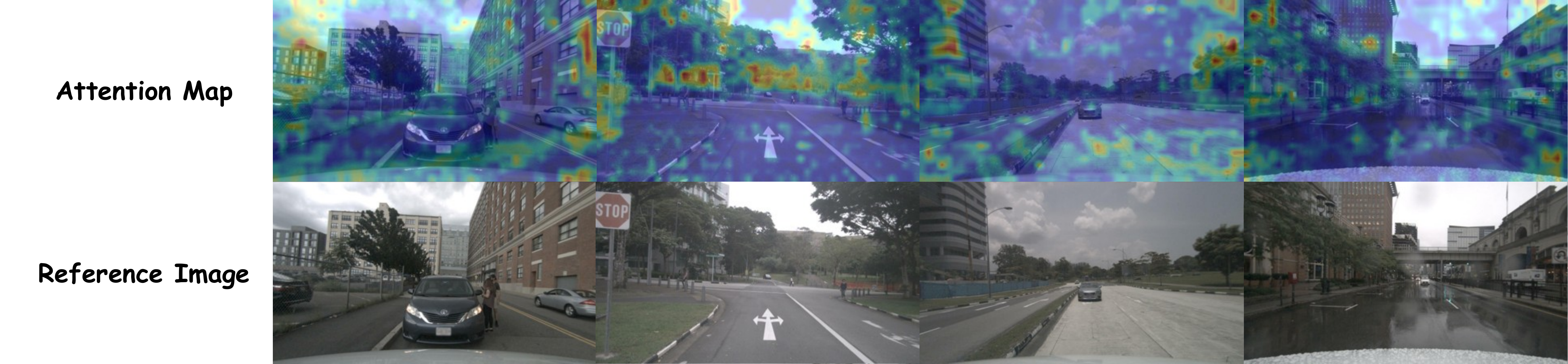}} 
	\caption{\textbf{Place-ID embedding contributed cross-attention visualization in generation process.}}
	\label{fig:6}
    \vspace{-20pt}
\end{figure}
Fig. \ref{fig:6} vividly demonstrates the contribution of place-ID embedding enables dynamic adjustment of attention toward background regions, thereby facilitating the integration of place-aware features.

\subsubsection{Training Support}
DiffPlace is capable of generating augmented data with precise annotations and consistent place-IDs, thereby bolstering the training process for autonomous vehicles' perception and place recognition tasks. For 3D object detection and place recognition, we augment the dataset with an equivalent number of images as in the original, maintaining consistent training iterations and batch sizes to ensure fair comparisons with the baseline. To optimize data augmentation, we randomly modify the weather to ``heavily rainy with wet road'' and alter half of the bounding boxes in each synthetic scene. 

\begin{table}[htbp]
    \vspace{-10pt}
	\caption{\textbf{Training Support for place recognition.} The results are reported on Pitts30k-test set.}
	\label{tab:2}
	\renewcommand \arraystretch {1.25}
	\centering
	\begin{tabular}{lllll}
		\toprule[1pt]
		\multirow{2}{*}{Data} & 
        \multicolumn{2}{c}{MixVPR} & 
        \multicolumn{2}{c}{CricaVPR} \\
		\cmidrule(lr{2pt}){2-3} \cmidrule(lr{2pt}){4-5}
		 & AR@1↑ & AR@5↑ & AR@1↑ & AR@5↑ \\
		\midrule
		w/o synthetic data & 83.5 & 90.3 & 90.9 & 96.0 \\ 
		MagicDrive & 84.2 & 91.1 & 90.3 & 95.7 \\ 
		Ours & \textbf{89.7} & \textbf{95.2} & \textbf{92.9} & \textbf{96.8} \\
		\bottomrule[1pt]
	\end{tabular}
    \vspace{-10pt}
\end{table}

\begin{figure}[htbp]
	\center{\includegraphics[width=9cm]  {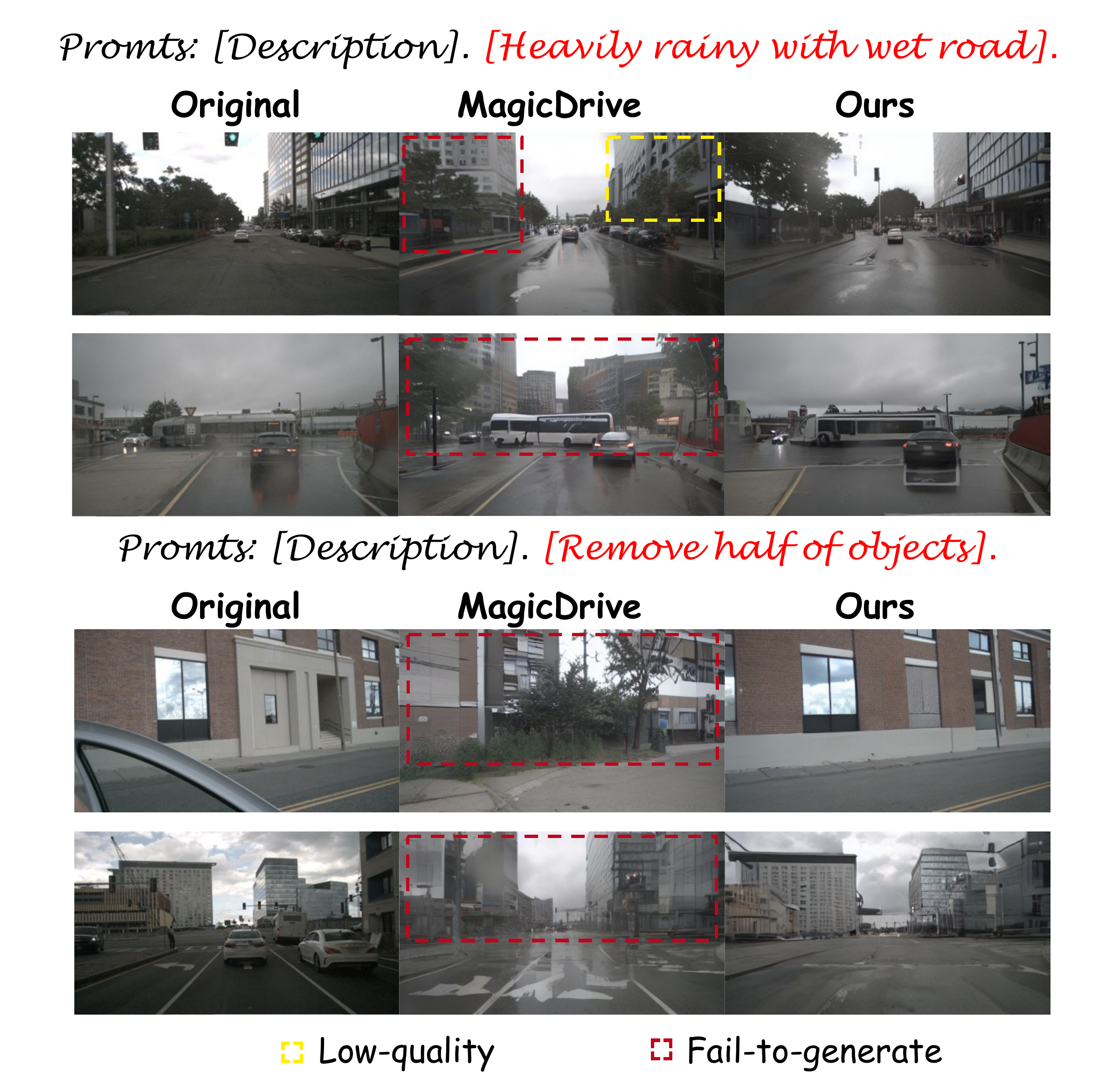}} 
	\caption{\textbf{Place controllablility under weather and objects edit.}}
	\label{fig:7}
    \vspace{-15pt}
\end{figure}

As shown in Fig. \ref{fig:7}, our method preserves the consistency of background buildings across variations in foreground objects and weather conditions. Table \ref{tab:2} highlights the beneficial impact of DiffPlace-generated data in training place recognition models including MixVPR \cite{MixVPR} and CricaVPR \cite{CricaVPR}, with AR@1 and AR@5 surpassing the baseline by 5.5\%, 4.1\% for MixVPR and 2.6\%, 1.1\% for CricaVPR, respectively. The inherent consistency in place-ID representation between the generated and original images, combined with increased data diversity, provides strong augmentation support for training place recognition networks.

\begin{figure}[htbp]
	\center{\includegraphics[width=\linewidth]  {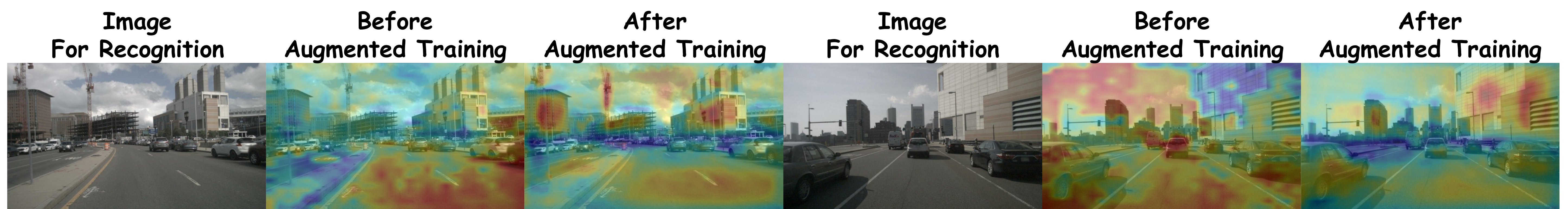}} 
	\caption{\textbf{Attention map extracted by CricaVPR on nuScenes test images.}}
	\label{fig:8}
    \vspace{-20pt}
\end{figure}

Moreover, we visualize the feature maps extracted by the state-of-the-art place recognition method CricaVPR on the images from nuScenes test set. As illustrated in Fig. \ref{fig:8}, after augmented training , the network shifts its attention towards background buildings, which are more reliable cues for place recognition. This demonstrates that the enhanced attention aligns better with semantic landmarks, indicating the effectiveness of our DiffPlace framework in improving place recognition models.

\begin{table}[htbp]
\vspace{-15pt}
	\caption{\textbf{Training Support for 3D Object Detection.} The results are reported on nuScenes validation set.}
	\label{tab:3}
	\renewcommand \arraystretch {1.25}
	\centering
	\begin{tabular}{lll}
		\hline
		\toprule[1pt]
		Data & mAP↑ & NDS↑ \\ \hline
		w/o synthetic data & 64.92  & 69.42 \\ 
		w/ MagicDrive & 67.28 & 70.14 \\ 
		Ours & 67.71 & 70.58 \\
		\bottomrule[1pt]
	\end{tabular}
    \vspace{-10pt}
\end{table}

Furthermore, as presented in Table \ref{tab:3}, DiffPlace achieves a marginal improvement over BEVFusion in the CAM+LiDAR setting \cite{BEVFusion}, which is competitive with MagicDrive. This demonstrates that our method enhances place recognition task without compromising its effectiveness in the original task.

\vspace{-5pt}
\subsection{Ablation Study}
As found in \cite{IP-Adapter, GPS2Pix}, we also find that using two linear projector can perform as well as cascaded  MLPs, while reducing computational complexity.
DiffPlace utilizes perceiver transformer to encode place-ID embedding. To demonstrate the efficacy, we train a model that directly takes the dimensional projection of place-ID embedding, denoted as “w/o perceiver transformer” in Table \ref{tab:4}.
Due to the limited information in place-ID embedding,  supplementary cross-attention mechanisms are crucial for enhancing the model's ability to precisely capture background information, as demonstrated by the disparity in recall rate performance.
Additionally, we evaluate the generation results without contrastive loss. In this case, without external force to make place conditions close to the CLIP space, the stability of training phase is reduced and it easily leads to bigger gap that reduces controllability.
\vspace{-5pt}
\begin{table}[htbp]
	\caption{\textbf{Ablation study on perceiver transformer in controller and contrastive learning strategy.} The results are recognition performance of MixVPR on synthesis images.}
	\label{tab:4}
	\renewcommand \arraystretch {1.25}
	\centering
	\begin{tabular}{lll}
		\hline
		\toprule[1pt]
		Data & AR@1↑ & AR@5↑ \\ \hline
		w/o perceiver transformer & 46.3 & 66.5 \\ 
		w/o contrastive loss & 51.1 & 70.2 \\
            cascaded MLPs & 57.2 & 75.1 \\
		Ours & \textbf{57.6} & \textbf{75.4} \\
		\bottomrule[1pt]
	\end{tabular}
    \vspace{-20pt}
\end{table}

\subsection{Limitations}

\begin{figure}[htbp]
	\center{\includegraphics[width=\linewidth]  {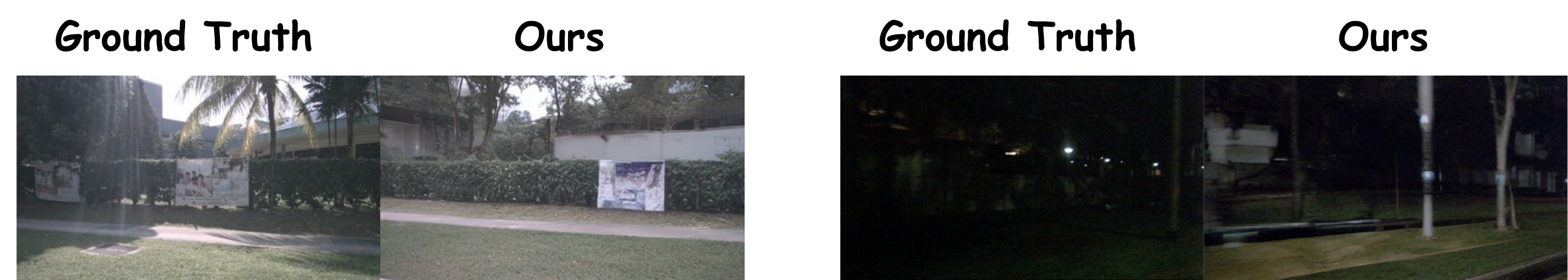}} 
	\caption{\textbf{Failure cases of DiffPlace.}}
	\label{fig:failure_case}
    \vspace{-15pt}
\end{figure}

We show representative failure cases in Figure \ref{fig:failure_case}. While DiffPlace generally performs well in generating driving scenes with consistent architectural backgrounds, it struggles in environments dominated by dense vegetation. This may be because the pre-trained place recognition models used to extract place-ID embeddings lack the fundamental ability to capture unstructured features such as foliage, making it more difficult for the generator to synthesize plant-heavy scenes.
Additionally, when the reference images are captured under extremely low-light conditions, the generated results often fail to match the ground truth. This can be attributed to two factors: (1) the pre-trained Diffusion model is not optimized for generating scenes under very dark lighting, and (2) the control signals, including the place-ID and reference image, do not provide sufficient information to guide faithful synthesis.
\vspace{-10pt}

\section{Conclusion}

In this work, we introduced DiffPlace, the first place-controllable diffusion model for generating augmented data to enhance place recognition performance. Extensive experiments indicate DiffPlace enables precious control over scene synthesis by place-ID controller, preserving background consistency across varying weather and foreground objects in generation. The augmented training experiments demonstrate that our method significantly enhances place recognition performance, yielding a 5.5\% improvement in AR@1 and a 4.1\% improvement in AR@5 for MixVPR, as well as gains of 2.6\% and 1.1\% for CircaVPR, respectively. These results highlight the effectiveness of DiffPlace in advancing place-aware street view generation and supporting robust place recognition. Future work may include training a street view video generative model with controllable background, midground and foreground on large-scale datasets.

% In this work, we presented DiffPlace, a novel framework for place-controllable multi-view image generation that significantly enhances the generated background features while maintaining other controllability. By introducing the place-ID controller, DiffPlace effectively addresses the limitations of existing methods, enabling the generation of urban scene images that maintain consistent background buildings while offering flexibility in modifying foreground objects and weather conditions. The proposed approach leverages place-ID encoding, a perceiver transformer, and contrastive learning to map these tokens into the CLIP latent space.

% Our extensive experiments demonstrate the superior place controllability of DiffPlace, which generates stree-view images with background that more closely resemble real-world buildings. This feature, coupled with the ability to produce varied foreground elements, proves essential for augmenting training data of place recognition tasks. The results in place recognition tasks further highlight the effectiveness of DiffPlace, with an impressive improvement of 5.5\% in AR@1 and 4.1\% in AR@5 over baseline method. These findings underscore the significant potential of DiffPlace in supporting the development of more effective generative models and more robust place recognition systems, particularly for autonomous driving applications.

\bibliographystyle{IEEEtran}
\bibliography{diffusion}

\end{document}